%% file: MAIN.tex
\documentclass[lettersize,journal]{IEEEtran}
\usepackage{amsmath,amsfonts}
\usepackage{algorithmic}
\usepackage{array}
\usepackage[caption=false,font=normalsize,labelfont=sf,textfont=sf]{subfig}
\usepackage{textcomp}
\usepackage{stfloats}
\usepackage{url}
\usepackage{verbatim}
\usepackage{graphicx}
\usepackage{amsmath}

\usepackage{nicefrac}

\usepackage{longtable}
\usepackage{booktabs}
\usepackage{multirow}
\usepackage{adjustbox}
\usepackage[accsupp]{axessibility} 
\usepackage{graphicx}
\usepackage{amsmath}
\usepackage{amssymb}
\usepackage{rotating}
\usepackage{caption}
\usepackage{booktabs}
\usepackage{paralist}
\usepackage{multirow}
\usepackage{color}
\usepackage{adjustbox}
\usepackage{cite}
\usepackage[accsupp]{axessibility} 
\usepackage{array}
\usepackage{makecell}

\usepackage[bottom]{footmisc}
\usepackage{nicefrac}
\usepackage{longtable}
\usepackage{booktabs}
\usepackage{scalerel}

\usepackage{multirow}
\newcolumntype{C}[1]{>{\centering\let\newline\small\arraybackslash\hspace{0pt}}m{#1}}

\usepackage{linguex}
\usepackage{tablefootnote}
\usepackage{makecell, multirow,
            tabularx} 
\usepackage{siunitx}
\usepackage{lipsum,afterpage,refcount}
\newcommand{\setfootnotemark}{%
  \refstepcounter{footnote}%
  \footnotemark[\value{footnote}]}
\usepackage{amsmath}
\usepackage[pagebackref,breaklinks,colorlinks]{hyperref}
\newcolumntype{R}[2]{%
    >{\adjustbox{angle=#1,lap=\width-(#2)}\bgroup}%
    l%
    <{\egroup}%
}

\usepackage[capitalize]{cleveref}
\crefname{section}{Sec.}{Secs.}
\Crefname{section}{Section}{Sections}
\Crefname{table}{Table}{Tables}
\crefname{table}{Tab.}{Tabs.}
\usepackage{balance}
\usepackage{titlesec}
\titlespacing*{\subsection}{0pt}{0.3\baselineskip}{0.4\baselineskip}

\titlespacing*{\section}{0pt}{0.6\baselineskip}{0.8\baselineskip}

\usepackage{caption}
\DeclareCaptionFont{ninept}{\fontsize{9.5pt}{11pt}\selectfont #1}

\usepackage{float}
\begin{document}
\title{Deep Pulse-Signal Magnification for remote heart rate estimation in compressed videos}

\author{\parbox{16cm}{\centering
{\large Joaquim Comas$^1$, Adrià Ruiz$^2$, Federico Sukno$^1$}\\
{\normalsize $^1$ Department of Information and Communication Technologies, Pompeu Fabra University, Barcelona, Spain \\ $^2$ Seedtag, Madrid, Spain}}} 


\maketitle

\maketitle
\input{Abstract/Abstract}
\input{Introduction/Introduction}
\input{Related_work/Related_work}
\input{Methodology/Methodology}

\input{Experiments/Experiments}

\input{Conclusions/Conclusions}

\input{Acknowledgment/Acknowledgment}

{\small
\bibliographystyle{abbrv}
\bibliography{MAIN}
}

\end{document}

%% file: Abstract/Abstract.tex
\begin{abstract}
Recent advancements in 
data-driven approaches for remote photoplethysmography (rPPG)
have significantly improved the accuracy 
of remote heart rate estimation. However, the performance of such 
approaches worsens considerably under video compression, which is nevertheless necessary to store and transmit video data efficiently. In this paper, we present a novel approach to address the impact of video compression on rPPG estimation, which leverages a pulse-signal magnification transformation to adapt compressed videos to an uncompressed data domain in which the rPPG signal is magnified. We validate the effectiveness of our model by exhaustive evaluations on two publicly available datasets, UCLA-rPPG and UBFC-rPPG, employing both intra- and cross-database performance at several compression rates. Additionally, we assess the robustness of our approach on two additional highly compressed and widely-used datasets, MAHNOB-HCI and COHFACE, which reveal outstanding heart rate estimation results.
\end{abstract}



%% file: Introduction/Introduction.tex
\section{Introduction}
\label{sec:intro}

Recently, the research community has shown increasing interest in the camera-based measurement of human physiological signals. These vital signs, including heart rate (HR), heart rate variability (HRV), respiration rate (RR), oxygen saturation (SpO2), and blood volume pulse (BVP) are essential for evaluating individuals' physical and mental state and have numerous potential applications \cite{ronca2021video, benezeth2018remote, liu20163d}. The advancements in multimedia technology have made many of these potential applications feasible, and have significantly impacted various industries such as education, entertainment, advertising, and healthcare. Thanks to current network connections, smartphones, and other multimedia devices, 
video content can be accessed or shared effortlessly. Nevertheless, in practical scenarios, the transmission of 
video content would not be viable without the use of video compression.

While significant progress has been made in remote physiological signal sensing, 
its application to compressed videos remains challenging. McDuff et al. \cite{mcduff2017impact} were among the first to demonstrate the impact of video compression, noting a significant reduction in the rPPG signal-to-noise ratio (SNR), even before human observers perceive a decrease in visual quality. Thus, it was found that even mild compression affects the recovery of the rPPG signal and, through this, the accuracy of derived estimates, such as HR or RR. 

Several other works \cite{hanfland2016video, cerina2019influence, nowara2019combating, gudi2020real} have 
later highlighted additional effects of video compression on automatic rPPG analysis, such as the effect of video resolution \cite{spetlik2018non}, the heterogeneous impact of compression for different color channels \cite{zhao2018novel}, or the greater impact of temporal compression with respect to chroma and spatial compression on the quality of the recovered rPPG signal \cite{rapczynski2019effects}. More details on these and other studies about the impact of video compression can be found in Section \ref{subsec:influence_compression}.



Unfortunately, to date, rather few solutions have 
addressed
the challenges 
of
video compression in rPPG estimation.
 They
can broadly be divided into three categories: 1) methods that propose to modify the recording conditions or the encoding algorithm \cite{zhao2018novel, zhao2019physiological}; 2) methods that propose to avoid a mismatch between the compression rates of training and test sets \cite{nowara2019combating,nowara2021systematic}; 3) methods 
that
try to reverse the damage introduced by the compressor \cite{yu2019remote} (details in Section \ref{subsec:rppg_solutions}).

In practice, the first category is of limited scope because it only applies to scenarios where videos are recorded specifically for rPPG estimation purposes. The second category is more widely applicable since it only requires finding a training dataset that matches the compression of the data being targeted; however, its success has been modest, yielding results that are still largely affected by the amount of compression \cite{nowara2019combating}. Thus, the most promising path so far has been the third category, where the most relevant approach was presented by Yu et al. \cite{yu2019remote}, based on an enhancement generator (STVEN) cascaded with a spatiotemporal convolutional network (rPPGNet) to recover the rPPG signal. The enhancement generator is framed within the idea of video quality enhancement (VQE) \cite{dong2015compression, zhang2017beyond}, which requires significant computational resources due to the large amount of data involved in restoring the high-frequency details lost due to video compression. 
Nevertheless, this is mainly due to the tendency of VQE methods to prioritize visual quality enhancement, which, as we will show, is not strictly needed to recover the rPPG signal.



\subsection{Contribution}
In this work, we propose a novel deep learning framework to mitigate the impact of video compression on rPPG recovery that departs from the objective of Visual Quality Enhancement and focuses, instead, on enhancing only the information required to allow proper recovery of the rPPG signal. To this end, we present a training methodology involving two deep neural networks: (i) The rPPG estimator network, trained on uncompressed videos, that attempts to recover the rPPG signal. (ii) The Pulse-Signal Magnification network (PSMN), that adapts compressed facial videos to a 
video domain in which the effects of the blood volume pulse are amplified, yielding a magnified pulse-signal and making possible the rPPG extraction 
even in highly compressed scenarios. 
We propose and validate a two-stage optimization procedure for both networks, in which the rPPG estimator acts as a regularizer to facilitate training of the PSMN, considerably improving its performance and yielding physiologically meaningful transformations.


We provide extensive experiments of
our approach, demonstrating its robustness and generalization through intra- and cross-dataset evaluation on 4 publicly available datasets: 2 uncompressed ones, UCLA-rPPG and UBFC-rPPG, and 2 highly compressed ones, MAHNOB-HCI and COHFACE.


The remainder of this paper is organized as follows: firstly, in Section \ref{sec:related}, we conduct a comprehensive review of the effects of video compression on rPPG estimation, the available solutions, and the prevalence of video compression in existing rPPG datasets. The proposed approach is presented in Section \ref{sec:method}, while experimental results are provided in Section \ref{sec:experiments}. Section \ref{sec:conclusions} summarizes our findings and conclusions.

%% file: Related_work/Related_work.tex
\section{Related work}
\label{sec:related}
\subsection{Camera-based Physiological measurement}

Since Takano et al. \cite{takano2007heart} and Verkruysse et al. \cite{verkruysse2008remote} showed the feasibility of measuring HR remotely from facial videos, many researchers have proposed different methods to recover physiological data. Among them, some works consider regions of interest using techniques such as Blind Source Separation \cite{poh2010non, poh2010advancements, lewandowska2011measuring}, Normalized Least Mean Squares \cite{li2014remote} or self-adaptive matrix completion \cite{tulyakov2016self}. In contrast, others 
rely on the skin optical reflection model by projecting all RGB skin pixel channels into an optimized subspace \cite{de2013robust, wang2015novel, wang2016algorithmic}. 

Later on,
deep learning methods \cite{vspetlik2018visual, yu2019remote, perepelkina2020hearttrack, lee2020meta, lu2021dual, nowara2021benefit} have outperformed conventional methods and achieved state-of-the-art performance in estimating vital signs from facial videos. Some of these methods leverage prior knowledge 
from traditional methods and combine it with Convolutional Neural Networks (CNNs) to exploit more sophisticated features \cite{niu2018synrhythm, niu2019rhythmnet, song2021pulsegan}. Some recent works \cite{lee2020meta, liu2021metaphys} have explored unsupervised approaches using meta-learning, showing improved generalization in out-of-distribution cases. On the other hand, 
other researchers have aimed at fully end-to-end approaches 
\cite{chen2018deepphys, Yu2019RemotePS, perepelkina2020hearttrack} 
 using facial videos 
as input to directly predict the rPPG signal. 
Attention-based models have recently gained traction in end-to-end approaches, with transformer-based models like Physformer \cite{yu2023physformer++} and EfficientPhys \cite{liu2023efficientphys} leading the way in leveraging long-range spatiotemporal features. Nevertheless, 
they have not yet demonstrated a significant performance advantage over CNN-based models \cite{liu2023efficientphys}. 


\subsection{Video-compression in existing rPPG datasets}
\label{subsec:video_compression}


Several datasets have been collected to advance the research on camera-based physiological sensing, addressing factors such as head movements, illumination conditions, and facial skin diversity. However, until recently, most of them were recorded in compressed video formats \cite{soleymani2011multimodal, zhang2016multimodal, heusch2017reproducible}, which has been found to negatively 
impact
the performance of rPPG systems (see Section \ref{subsec:influence_compression}). Thus, a few uncompressed datasets have recently emerged \cite{stricker2014non ,estepp2014recovering, bobbia2019unsupervised}, although not all of them are available to the public and have limited sample sizes, making it challenging to propose effective solutions, 
especially if they are based on deep learning.

Table \ref{table_datasets} summarizes existing rPPG datasets 
and highlights video compression details, including format, camera specifications, and approximate video quality 
in terms of bitrate \footnote{To collect the video information from the different datasets, we contacted the authors of each dataset and utilized the open-source FFmpeg ffprobe tool: https://ffmpeg.org/ffprobe.html}. Although we can distinguish different content formats in the existing rPPG datasets, the majority of them are stored using three standard compression codecs: MPEG-4 Video Part 2 (MPEG-4), Advanced Video Coding (H.264) and Motion JPEG (MJPG).

The most popular standards in Table \ref{table_datasets} are MPEG-4 and H.264, 
both based on block-wise motion compensation. H.264 is an improved version of MPEG-4, providing better compression due to variable block-size segmentation, optimal discrete cosine transform (DCT), and enhanced inter-frame prediction. While H.264 is the most common format, MPEG-4 is used in some datasets, including COHFACE \cite{heusch2017reproducible}, OBF \cite{li2018obf}, and CMU \cite{dasari2021evaluation}. VIPL-HR \cite{niu2018vipl} and UBFC-Phys \cite{sabour2021ubfc} use the MJPG codec, which compresses each frame individually using JPEG compression, minimizing inter-frame compression's impact on HR estimation \cite{niu2018vipl}. However, the JPEG 
quantization stage 
can also be challenging for rPPG solutions.

In contrast, uncompressed datasets have more diversity regarding codec format, with some storing raw data in image files using lossless formats such as PNG, BMP, or Bayer format (e.g., PURE \cite{stricker2014non}, or MR-NIRP \cite{magdalena2018sparseppg}), and others using uncompressed video formats, including YUV420, RV32, and RGBA format (e.g., ECG-Fitness \cite{vspetlik2018visual} or UBFC-rPPG \cite{bobbia2019unsupervised}). 

Regarding video quality, there are significant average bitrate differences 
between 
datasets. For example,
UCLA-rPPG \cite{wang2022synthetic} has 295 Mb/s 
while
COHFACE \cite{heusch2017reproducible} 
has just 0.250 Mb/s. In this case, both datasets use the same camera resolution, but the effects of video compression yield a notable difference in video quality. As noted by S\v{p}etl\'ik et al. \cite{spetlik2018non}, the encoding standard alone 
does not
determine video quality 
but other factors,
such as the pixel format or the camera resolution, must also be considered in the bitrate computation. For example, the VicarPPG and MAHNOB-HCI datasets use H.264 encoding but with different camera settings, resulting in highly different average bitrates. VicarPPG has a bitrate of 34 Mb/s, while MAHNOB-HCI has a bitrate of 4.20 Mb/s. This means that MAHNOB-HCI has approximately 8.5 times lower quality than VicarPPG.

\input{Related_work/table_datasets}

\subsection{Influence of video compression on rPPG estimation}
\label{subsec:influence_compression}

Video compression techniques optimize visual quality by removing information imperceptible to the human eye. However, the effects of heart beating on the skin are also imperceptible to the human eye and compression algorithms often 
remove subtle color variations that are important for inferring rPPG signals. 
Hanfland et al. \cite{hanfland2016video} were the first to show the influence of video compression on 
rPPG 
signal extraction. They compressed raw videos using Motion JPEG, MPEG-4, and Motion JPEG 2000 codecs and computed the time-domain correlation between the videos and extracted signals. 
A key conclusion from their work is that the rPPG signal is still present in the compressed videos, although they also find that 
its overall quality 
is altered by the compression. 

McDuff et al. \cite{mcduff2017impact} analyzed the impact of video compression on rPPG signal quality by testing two compression standards, H.264 and H.265, at different compression 
rates, 
They found noticeable degradation of the rPPG signal even under mild compression and concluded that only videos with a bit rate of at least 10 Mb/s maintained an rPPG signal with reasonable SNR. 
They also found H.265 to outperform H.264,
especially for videos with large movement variations. 
Similarly to McDuff et al., \cite{mcduff2017impact}, Cerina et al. \cite{cerina2019influence} analyzed the impact of frame rate and video compression on remote pulse-rate variability (PRV) using a lossless (FFV1) and lossy format (H.264) at two compression rates. They observed a notable trade-off between PRV quality and storage size in FFV1 format, while H.264 showed higher degradation in PRV, 
depending on peak detection accuracy. 

Other studies have further explored the impact of video encoding on camera physiological measurements. 
S\v{p}etl\'ik et al. \cite{spetlik2018non} identified errors in the acquisition stage and conducted an experiment similar to \cite{mcduff2017impact} to demonstrate the degradation of rPPG signal SNR at various compression rates using H.264 encoding. They found that lower resolutions had a higher impact on the recovered signal SNR and emphasized the negative effects of chroma subsampling, common in web camera hardware settings. 
Rapczynski et al.
\cite{rapczynski2019effects} analyzed the effects of different factors in video compression, finding that inter-frame compression was more damaging than intra-frame compression or chroma subsampling. Zhao et al. \cite{zhao2018novel} found that high compression levels affect blue and red color channels the most in terms of amplitude degradation, high-frequency noise, and trace discontinuity. Nowara et al. \cite{nowara2021systematic} extended the study of video compression on rPPG recovery to evaluate 
the effects of large head motion and skin color diversity.

Other works 
have
investigated the impact of video compression on the development of rPPG datasets. 
Niu et al. \cite{niu2018vipl} compressed the raw data using five different video codecs: MJPG, FMP4, DIVX, PIM1, and H264, and evaluated the accuracy of HR estimation. They 
found
that the MJPG codec provided a reasonable balance between data compression and preservation of rPPG signals. In contrast, Gudi et al. \cite{gudi2020real} 
explored the trade-off between rPPG accuracy and video size/bitrate in the PURE dataset. They 
tested several
lossy (VP9, H.264, H.265, MJPG, MPEG-4) and lossless (FFV1, HuffYUV) video codecs at 
multiple
bitrates. They concluded that H.265 and FFV1 were the most efficient codecs for preserving rPPG accuracy and storage, respectively.

Our findings, discussed in Section \ref{subsubsec:impact}, align with these prior studies that highlight the negative effects of video compression on the retrieval of rPPG signals. Moreover, in addition to validating these observations, we introduce a novel approach in Section \ref{sec:method} to mitigates the adverse effects induced by video compression. 

\subsection{rPPG solutions for compressed facial videos}
\label{subsec:rppg_solutions}

While several studies have evaluated the effects of compression on rPPG measurements, very few have proposed 
solutions 
to mitigate such effects.
Zhao et al. \cite{zhao2018novel} developed a single-channel pulse extraction method that only considers the green channel, as they found that it is 
the least affected one by
compression artefacts. However, ignoring the other channels leads to suboptimal estimations because important rPPG information is neglected. Another solution proposed by Zhao et al. \cite{zhao2019physiological} involved designing a new video codec that compresses regions not involved in rPPG extraction, but this approach is not practical for already compressed videos and is challenging to standardize across all multimedia content.

Data-driven solutions have also been explored. McDuff \cite{mcduff2018deep} introduced a deep super-resolution preprocessing step to restore high-frequency spatial details from low-resolution or intra-frame (spatial) compressed facial videos. Thus, this method does not consider the adverse effects of inter-frame (temporal) compression. In contrast, Nowara et al. \cite{nowara2019combating, nowara2021systematic} demonstrated that models trained on videos with the same compression level as the test set achieve better performance with respect to mismatched train-test compression; however, this 
requires knowledge of the compression rate of the testing videos and appropriately matched training data.

Finally, Yu et al. \cite{yu2019remote} presented the first enhancement method to deal with temporal compression using a video-to-video generator called STVEN and an rPPG estimator called rPPGNet. They employed a three-stage training process and multiple objective loss functions to optimize both models. The STVEN generator is firstly trained to reconstruct uncompressed videos from compressed ones and, independently, the rPPGNet estimator is optimized using high-quality videos. 
Finally, a joint training step is performed, fixing the parameters of the rPPGNet estimator and optimizing the STVEN generator to enhance facial videos for rPPG extraction.

In contrast to Yu et al., we propose a simpler two-stage approach that improves existing HR estimation in highly compressed scenarios by focusing solely on the physiological perspective. Our Pulse-Signal Magnification Network is designed to magnify rPPG characteristics in video rather than reconstructing all uncompressed video details, which are unnecessary for this task and make the model more difficult to train and less effective. 



%% file: Related_work/table_datasets.tex
\begin{table*}[h]
\renewcommand{\arraystretch}{1}
\centering
\footnotesize
\linespread{1}\selectfont\centering
\caption{Summary of the existing camera-based physiological signal sensing databases and its specifications. }
 \begin{tabular}{C{2.4cm} C{0.9cm} C{1.1cm} C{3.7cm} C{1.3cm} C{1.6cm} C{1.1cm} C{0.8cm} C{1.5cm} }
 \hline
 Dataset & Year &  Publicly available & Camera Settings & Average Bitrate & Data Format & Subjects  & Videos &  Signal acquisition \\[1.7ex] 
 \hline
MAHNOB-HCI \cite{soleymani2011multimodal} & 2011 & Yes & RGB camera, 780x580, 61 Hz  & 4.20 Mb/s & H.264 & 27 & 527 & ECG, GSR, RF, ST, EEG \\[1.5ex]

AFRL \cite{estepp2014recovering} & 2014 & No & RGB camera, 658x492, 30 Hz  & 310 Mb/s  & Raw (Bayer) & 25 & 300 & ECG, BVP \\ [1ex]

PURE \cite{stricker2014non} & 2014 & Yes & RGB camera, 640x480, 30 Hz & 222 Mb/s & Raw (PNG) & 10 & 59 &  PPG, SpO2  \\ [1.5ex]

VicarPPG \cite{tasli2014remote} & 2014 & Yes & Webcam, 720×1280, 30 Hz  & 34 Mb/s & H.264 & 10 & 20 & PPG  \\ [1ex]

BP4D \cite{zhang2016multimodal} & 2016 & No & Webcam, 1040x1392, 25 Hz & 3.5 Mb/s & H.264 & 140 & 1400 &  PPG, RF, HR, EDA \\ [1.5ex]
 
COHFACE \cite{heusch2017reproducible} & 2017 & Yes & Webcam, 640x480, 20 Hz & 0.25 Mb/s & MPEG-4 & 40 & 160 & PPG, RF \\ [1ex]
 
UBFC-rPPG \cite{bobbia2019unsupervised} & 2017 & Yes & Webcam, 640x480, 30 Hz & 215 Mb/s & Raw (RV24) & 42 & 42 &  HR, PPG \\ [1ex]

OBF \cite{li2018obf} & 2018 & No & RGB/NIR camera, 1920x1080, 60 Hz  & 20 Mb/s & MPEG-4 & 106 & 212 &  ECG, RF, PPG  \\ [1ex]
 
ECG-Fitness \cite{vspetlik2018visual} & 2018 & Yes & Webcam, 1920x1080, 30 Hz & 745 Mb/s & Raw (YUV420)  & 17 & 204 & ECG \\  [1.5ex]

MR-NIRP \cite{magdalena2018sparseppg} & 2018 & Yes & RGB/NIR camera, 640 × 640, 30 Hz & 590 Mb/s &  Raw (Bayer) & 18 & 37 & PPG  \\ [1ex]
 
PFF \cite{hsu2017deep} & 2020 & Yes & Webcam, 1280x720, 50Hz & 10 Mb/s & H.264 & 13 & 85 & HR \\ [1ex]
 
VIPL-HR \cite{niu2019rhythmnet} & 2020 & Yes & RGB/NIR camera, 960x720, 640x480, 25-30 Hz  & 5.15 Mb/s & MJPG & 107 & 3130 & HR, SpO2, PPG \\ [1ex]

MOLI-PPG \cite{perepelkina2020hearttrack} & 2020 & No & Webcam/RGB camera, 1280x720, 1920x720, 25-50 Hz  & Unknown & Raw (BMP) & 30 & 229 & PPG  \\ [1.7ex]

VicarPPG-2 \cite{gudi2020real} & 2020 & Yes & Webcam, 1280×720, 60 Hz  & 16.5 Mb/s & H.264 & 10 & 40 & PPG, ECG \\ [1ex]

CMU \cite{dasari2021evaluation} & 2021 & Yes & Smartphone, 25×25, 15 Hz & 0.145 Mb/s & MPEG-4 & 140 & 140 &  HR \\ [1ex]

CameraHRV \cite{pai2021hrvcam} & 2021  & Yes & RGB camera, 1920×1200, 30 Hz & 1659 Mb/s & Raw (Bayer) & 14 & 60 & PPG  \\ [1ex]
 
UBFC-Phys \cite{sabour2021ubfc} & 2021 & Yes & RGB camera, 1024×1024, 35 Hz  & 225 Mb/s & MJPG & 56 & 168 &  PPG, EDA \\ [1ex]

DDPM \cite{speth2021deception} & 2021 & Yes & RGB/NIR camera, 1920 × 1080, 90 Hz & 950 Mb/s & H.264 \setfootnotemark\label{first} & 86 & 86 & PPG, SpO2, HR \\ [1ex]

UCLA-rPPG \cite{wang2022synthetic} & 2022 & Yes & RGB camera, 640x480, 30 Hz  & 295 Mb/s & Raw (RGBA) & 98 & 489 & PPG, HR \\ [1ex]

 \hline
 \end{tabular}
 \label{table_datasets}
 \afterpage{\footnotetext[\getrefnumber{first}]{The original camera pixel format was YUV422P format, but it was compressed in a lossless H.264 using a Constant Rate Factor (CRF) of 0}}
 \end{table*}

%% file: Methodology/Methodology.tex
\section{Methodology}
\label{sec:method}
In this section, we introduce our proposed model, which is composed of the rPPG estimator and the Pulse-Signal Magnification network. Subsequently, we detail our two-stage training procedure and present our optimized objective function for rPPG recovery.

\subsection{Problem formulation}

Our framework aims to extract the rPPG signal from compressed videos by 
learning a suitable transformation of the video domain. This transformation aims to magnify 
the tiny pulse signal that is present in the video but has possibly been damaged or distorted by the compression algorithm. A key element for the success of our approach is to focus on restoring only the part of the video signal that is necessary to the rPPG estimation, but not the video as a whole, which would lead to a considerably more complex system.

Let $\mathbf{C^n}=[c^n_1,c^n_2,...,c^n_T]$ be a facial compressed video $n \in N_c$, where $N_c$ is the total number of compressed videos with the number of frames $t \in T$ in our training set. The corresponding rPPG ground-truth signal from the compressed $\mathbf{C^n}$ is denoted as $\mathbf{y_c}^n=[y_{c,1}^n,y_{c,2}^n,...,y_{c,T}^n] \in \mathcal{Y}_c$ and it represents the pulse signal that we wish to extract from the video.

Given the above definitions, we denote our compressed rPPG video dataset as $\mathcal{D^C}=[(\mathbf{C}^1,\mathbf{y_c}^1),...,(\mathbf{C}^{N_c},\mathbf{y_c}^{N_c})]$. Similarly, we assume that, during training, we have also access to 
another set of facial videos without compression\footnote{It is also possible to use the same videos in set $\mathcal{D^C}$ before compression, but this is not required.},
$\mathbf{U}^n=[u^n_1,u^n_2,...,u^n_T] \in \mathcal{U}$ with $n \in N_u$, where $N_u$ is the total number of uncompressed videos, also with their corresponding PPG ground-truth $\mathbf{y_u}^n=[y_{u,1}^n,y_{u,2}^n,...,y_{u,T}^n]$, forming 
the rPPG uncompressed dataset $\mathcal{D^U}=[(\mathbf{U}^1,\mathbf{y_u}^1),...,(\mathbf{U}^{N_u},\mathbf{y_u}^{N_u})]$.

As previously mentioned, our two-stage framework involves training two neural networks: $f_\theta$ and $m_\psi$. The network $f_\theta$ learns the mapping from uncompressed facial videos $\mathcal{U}$ to rPPG signal $\mathcal{Y}$, while the network $m_\psi$ maps the compressed RGB facial videos $\mathcal{C}$ to a new pulse-signal magnification subspace $\mathcal{M}$:
\begin{equation}
  \label{eq:u}
  \begin{gathered}
     f_\theta : \mathcal{U} \to \mathcal{Y}\\        
     m_\psi : \mathcal{C} \to \mathcal{M} 
  \end{gathered}
\end{equation}
where $\theta$ and $\psi$ refer to the rPPG and pulse-magnification network, respectively.
Our objective is to partially approximate the relationship between uncompressed $\mathcal{U}$ and compressed $\mathcal{C}$ facial videos by learning a video transformation that effectively amplifies the information related to the pulse signal that was distorted by the compression effect:

\begin{equation}
    \mathbf{\hat{y}_M}^n =  f_\theta(m_\psi(\mathbf{C}^n)) 
\end{equation}

\noindent where 
$\mathbf{\hat{y}_M}^n$
represents 
the rPPG signal estimated from pulse-magnified video, i.e. from the compressed facial video $\mathbf{C^n}$ once it has been conveniently transformed 
into $m_\psi(\mathbf{C}^n) \in \mathcal{M}$. Notice that, in general, $\mathcal{M} \neq \mathcal{U}$ except from the perspective
of the recovered signal, $\mathbf{\hat{y}_M}^n$.

\begin{figure*}
    \centering
\includegraphics[width=165mm,height=80mm]{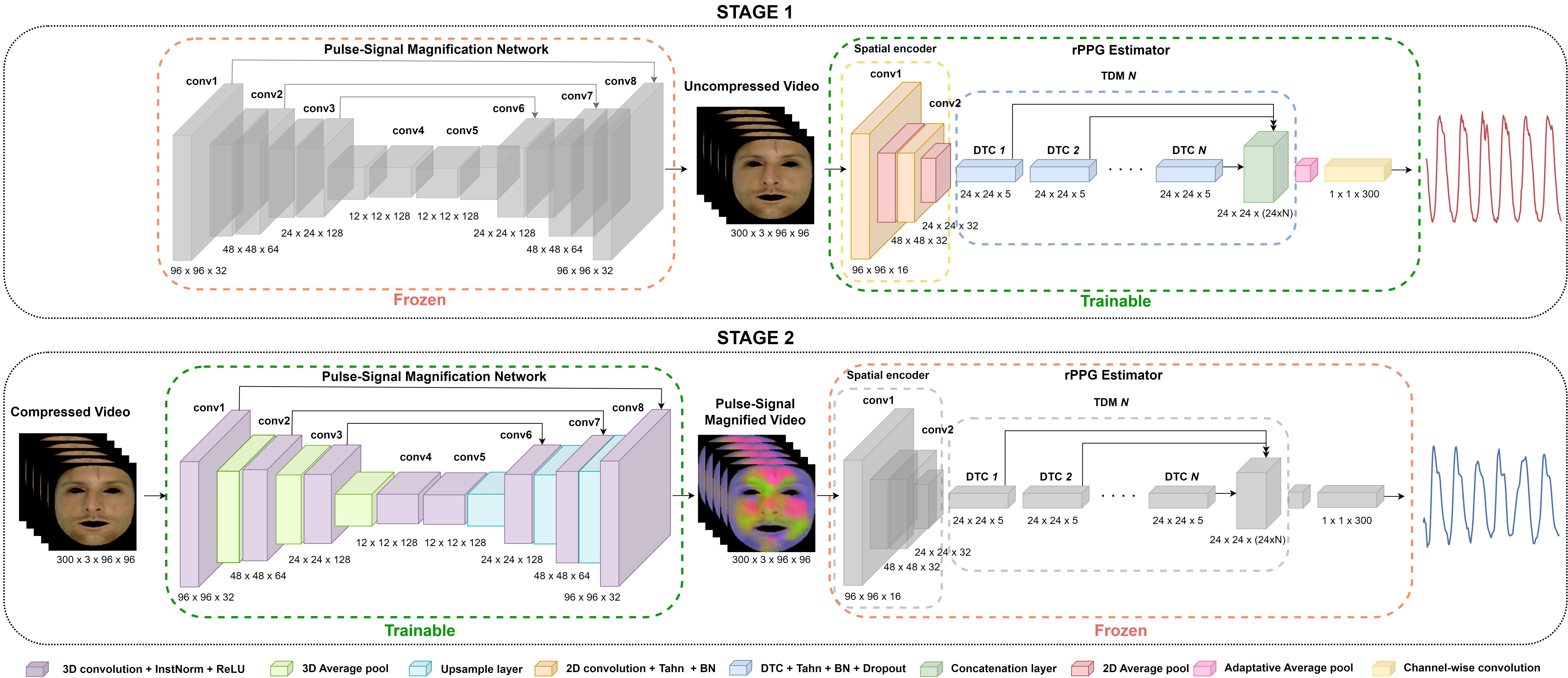}
    \caption{Overall structure of our proposed model, which has two stages for rPPG recovery under compression. Firstly, the TDM model is trained on uncompressed data. Then, the Pulse-Signal Magnification network is trained on compressed data with fixed TDM model parameters.}
    \label{fig:model}
\end{figure*}

\subsection{Networks Architecture}

An illustration of the two networks previously described is provided in Figure \ref{fig:model}. The rPPG estimator network $f_\theta$ is a lightweight spatiotemporal model presented in \cite{comas2022efficient}. This network effectively estimates rPPG spatiotemporal features by aggregating temporal derivative modules (TDM), emulating a Taylor series expansion. We selected this architecture because it achieves competitive results in current benchmarks while using fewer parameters than 
other alternatives with comparable performance.

The Pulse-Signal Magnification network $m_\psi$ is an Unet-based network that learns the desired pulse-signal magnification transformation. This spatiotemporal network comprises three 3D convolutions with a 3x3x3 kernel, followed by ReLU activation functions, 3D instance normalization, and average pooling. We then employ two spatiotemporal convolutions in the lowest dimensionality space and reconstruct the latent space using three 3D upsampling convolutions. At each resolution level, we incorporate a skip connection to 
lead the network to produce the same facial data as the input.

\subsection{Two-stage Training Framework}
The motivation for our proposed two-stage training framework is three-fold: 1) to better handle the domain gap between compressed and uncompressed videos in the recovery of remote PPG signal by magnifying the pulse signal in the video domain, 2) to leverage a pre-trained rPPG estimator network $f_\theta$ on an uncompressed video dataset as a regularizer for the Pulse-Signal Magnification network $m_\psi$, and 3) to reduce the training complexity of the overall system. 
By decoupling the training into two stages, we can ensure that the Pulse-signal Magnification network $m_\psi$ learns a representation that better approximates the performance of uncompressed videos. 

\subsubsection{Stage I} The first stage of the framework consists of training our $f_\theta$ as rPPG estimator from uncompressed data $\mathcal{D^U}$. This stage aims to obtain a pre-trained network that can estimate the PPG signal accurately from uncompressed videos, which can be used as a regularizer for the Pulse-Signal Magnification network $m_\psi$ in the second stage. This stage is essential in addressing the domain gap between compressed and uncompressed videos, as the pre-trained rPPG estimator network $f_\theta$ can capture the signal from uncompressed videos more effectively than using compressed data. Therefore, the objective of this first stage is to optimize the $f_\theta$ network over $\theta$ parameters:
\begin{equation}
    \min_\theta \frac{1}{N_u} \sum_{n=1}^{N_u} \mathcal{L}_\mathrm{{rPPG}}(f_\theta(\mathbf{U}^n),\mathbf{y_u}^n),
\end{equation}
\noindent where $n$ is the index of the uncompressed video $\mathbf{U}^n$, $\mathbf{y_u}^n$ the corresponding PPG ground-truth signal, and $\mathcal{L}_\mathrm{{rPPG}}$ is the rPPG loss function that measures the difference between the estimated PPG signal and the PPG ground-truth signal, explained in Section \ref{subsec:loss_function}. 
\subsubsection{Stage II}
\label{subsec:stage2}
The second stage of the framework involves training a Pulse-Signal Magnification network $m_\psi$ to learn 
an optimal video transformation. In a study by Nowara et al. \cite{nowara2019combating}, it was argued that training deep learning models with the same compression level as the testing data yields the best performance. Following this idea, the pre-trained rPPG estimator model $f_\theta$ is designed to estimate the rPPG signal from non-compressed videos, while the Pulse-Signal Magnification network $m_\psi$ learns to map videos in the domain of $\mathcal{C}$ to a domain that magnifies the pulse-signal from compressed data $\mathcal{D^C}$ so that it behaves like non-compressed data $\mathcal{D^U}$
. In this way, it adapts the compression
level of input data to the uncompressed format guided by the knowledge stored in the $f_\theta$ network at Stage I.

Thus, the objective of the second stage is to jointly train both networks such that the output of $m_{\psi}$ is regularized by $f_\theta$. The parameters $\theta$ of the rPPG estimator network $f_\theta$ are kept frozen during this training stage, i.e. only the parameters $\psi$ of the Pulse-Signal Magnification network $m_\psi$ are updated:
\begin{equation}
    \min_\psi \frac{1}{N_c} \sum_{n=1}^{N_c} \mathcal{L}_\mathrm{{rPPG}}(f_\theta(m_\psi(\mathbf{C}^n)),\mathbf{y_c}^n),
\end{equation}
\noindent where $n$ is the index of the compressed video $\mathbf{C}^n$, $\mathbf{y_c}^n$, is the corresponding ground-truth PPG signal, and $\mathcal{L}_{rPPG}$ is the rPPG loss function. In Section \ref{subsubsec:compression_unet_ablation}, we will discuss the advantages of this two-step approach and present empirical evidence of its effectiveness in our experimental section.

\subsection{Loss function}
\label{subsec:loss_function}
We 
use a combined loss function focused solely on the physiological aspect to 
guide 
the optimization of our model toward improving the pulse signal from the input video. The combined loss function comprises a temporal loss function aimed at restoring the intrinsic characteristics of the PPG waveform, and a frequency loss function, aimed at learning spectral features from the HR distribution.

For the temporal loss function, we adopt TALOS (Temporal Adaptive LOcation Shift) Loss, presented in \cite{comas2022efficient} that allows training of deep learning methods invariantly to temporal offsets of the ground-truth signal, also known as Pulse Transit Time (PTT). Therefore, our $\mathcal{L}_\mathrm{{temp}}$ can be written as:
\begin{equation} \label{eq:TALOS}
    \mathcal{L}_\mathrm{{temp}} = \sum_{k\in K} MSE(\hat{y_t}, {y}_{(t-k)}) \cdot p_{\mathbf{\theta}}(k|s),
\end{equation}
\noindent where the mean square error (MSE) is computed between the predicted rPPG signal $\hat{y_t}$ and the ground-truth PPG signal $y_{(t-k)}$ for each possible offset k and is weighted for each subject $s$ according to the learned temporal-shift probability $p_{\mathbf{\theta}}(k|s)$.

For the frequency loss function, we employ the signal-to-noise ratio (SNR) loss \cite{de2013robust} to incorporate the HR estimation task as a classification problem in the frequency domain. Therefore, our  $\mathcal{L}_\mathrm{{freq}}$ can be written as:
\begin{equation}
    \mathcal{L}_\mathrm{{freq}} = CE(PSD(\hat{y_t}), HR_{gt}),
\end{equation}
\noindent where $CE$ is the cross-entropy loss between the power spectral density $PSD(\hat{y_t})$ of the predicted rPPG signal $\hat{y}$ and the HR ground-truth value, $HR_{gt}$.

In summary, the overall loss function $\mathcal{L}_\mathrm{{rPPG}}$ for both networks can be formulated as:
\begin{equation}
\label{combined_eq}
    \mathcal{L}_\mathrm{{rPPG}} = \mathcal{L}_\mathrm{{temp}} + \lambda \cdot \mathcal{L}_\mathrm{{freq}},
\end{equation}
 where $\lambda$ is a balancing parameter. In our experiments, we set $\lambda = 0.01$ empirically based on our preliminary experiments.

\subsection{Constrained Pulse-Signal Magnification space}
\label{subsec:newspace}

Once our loss is formalized, let us revisit the concept behind our training stage II, introduced in Section \ref{subsec:stage2}. While the problem was initially 
described similarly to a video quality enhancement task,
we do not actually target the restoration of the compressed video to obtain its uncompressed form, since there are no video reconstruction terms in our loss function. Instead, we only target the minimization of the estimated rPPG signal and the HR extracted from it, respectively, with $\mathcal{L}_\mathrm{{temp}}$ and $\mathcal{L}_\mathrm{{freq}}$. This does not lead to the enhancement of the visual quality of the transformed video, but rather to its transformation into an alternative, arbitrary domain $\mathcal{M} \neq \mathcal{U}$ 
that is determined based on the aforementioned losses.

Thus, $m_\psi(\mathbf{C}^n)$ is not a decompressed version of $U^n$, but 
the result of a
transformation into a latent physiological domain in which the rPPG signal can be extracted with an accuracy similar to the uncompressed case, yet while starting from the compressed video input. As we shall see in the experiments, the transformed video is largely distorted in terms of visual quality, but it effectively focuses on the intensity changes that are relevant to estimate the rPPG signal. Indeed, after processing the input video with our Pulse-Signal Magnification network $m_\psi$, the pulsatile effect on the skin becomes magnified, to the extent that it can be appreciated by the naked eye, motivating the name of the method. This pulse-signal magnification space will be further analyzed in Section \ref{subsubsec:compression_unet_ablation}.



%% file: Experiments/Experiments.tex
\section{Experiments}
\label{sec:experiments}


In this section, we present the 4 benchmark datasets used in our experiments and describe the implementation and results of our method. We begin by examining the effect of the loss function, followed by investigating the impact of video compression on rPPG recovery using our baseline model.
Next, we evaluate the performance of our two-stage model through intra-database and cross-database evaluations, showing its robustness to video compression. Finally, we present a 
comparison with existing rPPG approaches on compressed benchmarks.

\subsection{Datasets}
\label{subsec:datasets}
We evaluate our approach on the following RGB video datasets.

The \textbf{UCLA-rPPG} dataset \cite{wang2022synthetic} comprises 489 videos from 98 subjects with diverse characteristics, including skin tone, ages, gender, and ethnicity. 
Each subject underwent 5 trials, with each trial lasting approximately 1 minute. The recordings were captured at a resolution of 640 x 480 pixels and 30 frames per second (FPS), in an uncompressed format, with an average bit rate of 295 Mb/s. Synchronous gold-standard PPG and HR measurements 
are provided alongside the facial videos.
Due to the lack of predefined 
evaluation protocol
in this dataset, we split the data ourselves into training (80\%), validation (10\%), and testing (10\%) sets.

The \textbf{UBFC-rPPG} dataset \cite{bobbia2019unsupervised} includes 42 RGB videos from 42 subjects. 
The recorded facial videos were acquired indoors with varying sunlight and indoor illumination at 30 FPS with 
a resolution of 640x480 in uncompressed 8-bit RGB format, with an average bit rate $\approx$ 215 Mb/s. 
PPG signal and heart rate 
at a sampling rate of 30 Hz are also provided.
In our experiments, we use UBFC-rPPG in a cross-dataset evaluation, where all 42 videos are used for testing.


The \textbf{COHFACE} dataset \cite{heusch2017reproducible} consists of 160 videos from 40 subjects, where each subject was recorded in 4 trials captured under two different lighting environments: studio and natural light. Each video was recorded 
at 20 FPS with a resolution of 640x480 for 1 minute and stored with heavy MPEG-4 compression, with an average bit rate $\approx$ 0.25 Mb/s. The physiological data 
consists of PPG and respiration ground-truth signals with a sampling rate of 256 Hz. For our comparison, we follow the preassigned standard folds defined in the evaluation protocol for the dataset.

The \textbf{MAHNOB-HCI} dataset \cite{soleymani2011multimodal} 
includes 527 facial videos from 27 participants, 
with corresponding physiological signals. The videos were recorded at a resolution of 780x580 and 61 FPS, 
and were heavily compressed using the H.264/MPEG-4 AVC H.264/MPEG-4 AVC codec, with an average bit rate $\approx$ 4.20 Mb/s. 
Each recording contains 
several biosignals, from which we use the electrocardiogram (ECG) to extract the HR ground truth with 
the Bob Toolbox\footnote{\url{https://www.idiap.ch/software/bob/}} \cite{heusch2017reproducible}. To compare fairly with previous works \cite{ yu2019remote, chen2018deepphys,lee2020meta}, we use the standard 30-second clip (frames 306 to 2135) of each video.
\subsection{Implementation details}
\label{subsec:setup}

\subsubsection{Preprocessing and training}
We adopt the same preprocessing stage for each dataset in all our experiments. Firstly, 
we apply a facial 
segmentation to remove the background and non-skin areas adapting the Mediapipe Face Mesh\footnote{\url{https://github.com/google/mediapipe/blob/master/docs/solutions/face_mesh.md}} model \cite{kartynnik2019real}. This step reduces the complexity of video compression by focusing attention on the facial skin without introducing more complexity in terms of parameters or optimization tasks. Once the facial video is masked, each video frame is resized to $96\times 96$ pixels. 
The ground-truth bio-signal is preprocessed following \cite{dall2020prediction} to denoise the raw PPG signal, which facilitates a better model convergence during 
training.

We implement our model using Pytorch 1.8.0 \cite{paszke2019pytorch} and train it on a single NVIDIA GTX1080Ti. We use sequences of 300 frames with an overlap of 10 
frames and use Adam optimizer 
with a learning rate of 0.0001 and weight decay of 1e-5. In addition, when using 
the
TALOS loss, we incorporate an extra SGD optimizer with a learning rate of $0.01$ to optimize the parameters $\mathbf{\theta}^s$ of the temporal-shift distributions 
in Eq. (\ref{eq:TALOS}).
Finally, the estimated HR is computed from the predicted rPPG signal using the power spectral density (PSD). Before calculating the HR value, we apply a band-pass filter with cutoff frequencies of 0.66 Hz and 3 Hz.

\subsubsection{Video compression settings}
\label{subsec:setup_compression}

To evaluate the impact of video compression
, we adopt the methodology from \cite{mcduff2017impact} 
for our ablation studies. 
Specifically, we produce compressed versions of the original uncompressed datasets by applying the H.264 codec with FFmpeg at various compression rates, categorized by the Constant Rate Factor (CRF). The CRF is a parameter that
maintains a constant visual quality level by dynamically adjusting the bitrate based on video content complexity 
and ranges from 0 to 51, with lower values indicating higher quality. For our ablation studies, we consider CRF values of 0, 5, 10, 15, 20, and 25, with 0 representing the uncompressed data.

\subsubsection{Metrics and evaluation}      
To evaluate the HR estimation performance 
we adopt 
the mean absolute HR error (MAE), the root mean squared
HR error (RMSE) and Pearson’s correlation coefficients R, widely used in the literature \cite{li2014remote, liu2023efficientphys}. 
For the UCLA-rPPG intra-dataset experiments, we use 300-frame sequences (10-second 
windows) without overlap for HR estimation. This approach is more challenging and informative compared to estimating HR based on the entire video sequence at once. For the remaining datasets, we compute whole-video performance 
given that most prior work follows that strategy.
This allows for a direct comparison of our approach to several traditional and deep learning methods.

\subsection{Ablation studies}
\label{subsec:ablation}

In Section \ref{subsec:video_compression}, we emphasized the importance of utilizing an uncompressed dataset for optimal results in our two-stage framework. As a result, we primarily choose the publicly available UCLA-rPPG dataset as the training dataset for our ablation studies. This dataset meets our criteria by including a substantial number of subjects with diverse skin types and being in an uncompressed format.

In this section, we present the results of our ablation studies on HR estimation. We conduct these studies using part of the UCLA-rPPG dataset for intra-dataset evaluation and the UBFC-rPPG dataset for cross-dataset evaluation. Firstly, we inspect the impact of the loss function on HR estimation using our baseline model. Next, we examine the effect of video encoding by applying different CRF values to recover the rPPG signal within the baseline model. Finally, we evaluate the performance of our proposed training procedure to mitigate the degradation introduced by video compression.
\\

\subsubsection{\textbf{Impact of loss function}}


In this initial experiment, we evaluate the performance of 
the selected rPPG baseline, the TDM model \cite{comas2022efficient},
using different loss functions: TALOS as a temporal loss function, SNR loss as a frequency loss function, and the proposed combination of both, from Eq. \ref{combined_eq}. Table \ref{table:loss_ablation} presents the HR results of the TDM baseline for each loss function in both intra- and cross-dataset evaluations.

Based on the results, we notice that the TDM model yields better HR results when trained with the TALOS loss function ($\mathcal{L}_\mathrm{{temp}}$) as opposed to the SNR loss. However, during cross-dataset evaluation, we find that the SNR loss exhibits better generalization capabilities compared to the temporal loss. These outcomes align with the findings of Yu \emph{et al.} \cite{yu2023physformer++}, who highlight the challenges associated with training rPPG approaches using either temporal or spectral losses exclusively. The temporal loss offers advantages in extracting signal trend features, but it carries the risk of overfitting and limited generalization. Conversely, the SNR loss facilitates the learning of finer periodic features by leveraging the HR frequency bands 
but
the presence of noise and the complexity of PPG waveforms can 
hamper the convergence of the model. Finally, we note that the combination of both losses produces a better performance than each one individually, similarly to \cite{yu2020autohr,yu2023physformer++}. In the intra-database results, we appreciate almost the same performance combining both losses with respect to TALOS loss while in the cross-dataset evaluation, we observe a significant reduction of both MAE and RMSE compared with each of the losses individually, indicating better generalization ability. Thus, these results support our choice of the combined loss presented in Eq. \ref{combined_eq}.
\\

\begin{table}
\centering
\caption{Impact of the loss function in HR measurement in intra-dataset and cross-dataset evaluation.}
\renewcommand{\arraystretch}{1.35}
\centering
\adjustbox{width=0.49\textwidth}{
  \begin{tabular}{c|c c c|c c c}
    \hline
    \multirow{3}{0.8cm}{Loss function} & \multicolumn{3}{|c}{Intra-dataset evaluation } & \multicolumn{3}{|c}{Cross-dataset evaluation}\\\cline{2-7}
    \multirow{3}{0.8cm}{ } & \multicolumn{3}{|c}{UCLA-rPPG} & \multicolumn{3}{|c}{UBFC-rPPG}\\ 
    \cline{2-7}
    & MAE$\downarrow$&RMSE$\downarrow$&R$\uparrow$
    & MAE$\downarrow$&RMSE$\downarrow$&R$\uparrow$
    \\ 
    \hline
    $\mathcal{L}_\mathrm{{temp}}$  & 1.00 & 4.43 & 0.90 & 2.54 & 6.18 & 0.93 \\[0.2ex]
    $\mathcal{L}_\mathrm{{freq}}$ & 1.16 & 4.51 & 0.89  & 2.38 & 5.91& 0.94 \\ [0.2ex]
    $\mathcal{L}_\mathrm{{temp}} + \mathcal{L}_\mathrm{{freq}}$   & 1.02 & 4.45 & 0.90  & 1.66 & 5.54 & 0.95  \\ [0.2ex]
    \hline
    \end{tabular}
    }
    \label{table:loss_ablation}
    \end{table}	
    						
\subsubsection{\textbf{ Impact of video compression in HR estimation}}
\label{subsubsec:impact}

In this section, we study the impact of the H264 encoding in the recovery of the rPPG signal using the TDM baseline model. To this end, we repeat the experiments from the previous section (keeping only the combined loss) but using increasingly compressed versions of the UCLA-rPPG and UBFC-rPPG datasets. Compression is applied to the whole dataset so that there is no compression mismatch between training and test sets. Table \ref{table:derivative_impact} summarizes the HR errors for different compression levels in terms of the CRF values. 

For the intra-dataset evaluation, we observe a gradual increase in HR error as the compression rate increases. However, for CRFs between of 5 and 15, the HR error shows a rather minor increase with respect to the uncompressed baseline. Beyond a CRF of 20, we observe a significant rise in HR error. Such CRF values correspond to about 450 kb/s, which is 4 times lower than the bit rate at CRF 15 and about 655 times lower than the uncompressed video. The observed trend coincides with previous findings, which 
reported significant degradation of the rPPG signal beyond this CRF value, especially when the facial videos are downsampled \cite{spetlik2018non}.

In contrast to the intra-database evaluation, the cross-dataset evaluation shows a substantial increase in HR error, even for the smallest CRF levels. This indicates that, while the TDM baseline performs well with medium compression levels within the same database, it struggles to generalize effectively to different types of data, even when using the same compression level as the training set. This limitation can be attributed to the TDM model primarily capturing intrinsic temporal cues of rPPG through temporal derivatives, which are more sensitive to compression artifacts 
under mismatched conditions of the training and test data,
such as motion or illumination. Nevertheless, to the best of our knowledge, we are the first to report 
the effect of different compression levels in cross-dataset experiments, hence this limitation might be shared by other data-driven models and not be exclusive of the TDM baseline.
\\


\begin{table}
\renewcommand{\arraystretch}{0.95}
\centering
\caption{Impact of video compression for rPPG recovery in intra-dataset and cross-dataset evaluation.}
\renewcommand{\arraystretch}{1.3}
\centering
\adjustbox{width=0.49\textwidth}{
  \begin{tabular}{c|c c c|c c c}
    \hline
    \multirow{3}{1cm}{CRF} & \multicolumn{3}{|c}{Intra-dataset evaluation } & \multicolumn{3}{|c}{Cross-dataset evaluation}\\\cline{2-7}
    \multirow{3}{1cm}{ } & \multicolumn{3}{|c}{UCLA-rPPG} & \multicolumn{3}{|c}{UBFC-rPPG}\\ 
    \cline{2-7}
    & MAE$\downarrow$&RMSE$\downarrow$&R$\uparrow$  
    & MAE$\downarrow$&RMSE$\downarrow$&R$\uparrow$
    \\ 
    \hline
    0  & 1.02 & 4.45 & 0.90  & 1.66 & 5.54 & 0.95   \\
    5  & 1.12  & 4.52 & 0.90 & 3.14 & 7.26 & 0.91   \\
    10 & 1.18 & 4.58 & 0.89 & 4.57 & 12.33 & 0.77       \\ 
    15 & 1.26 & 4.69 & 0.89 & 6.75 & 14.74 & 0.71  \\
    20 & 1.94 & 7.25 & 0.76 & 12.82 & 21.83 & 0.49  \\ 
    25 & 2.98 & 9.33 & 0.72 & 14.44 & 24.5 & 0.43 \\ 
    \hline
    \end{tabular}
    }
    
    \label{table:derivative_impact}
    \end{table}

\subsubsection{\textbf{Impact of network training procedure}}
\label{subsubsec:compression_unet_ablation}


We conduct an evaluation of the training procedure using two different strategies: $i)$ the proposed two-stage training, and $ii)$ end-to-end training. Differently from the previous two sections, in which only the TDM module was used, we now incorporate also the Pulse-Signal Magnification network, thus deploying our full system.

\begin{figure}[h]
\centering
    \includegraphics[width=3.5cm,height=9cm]{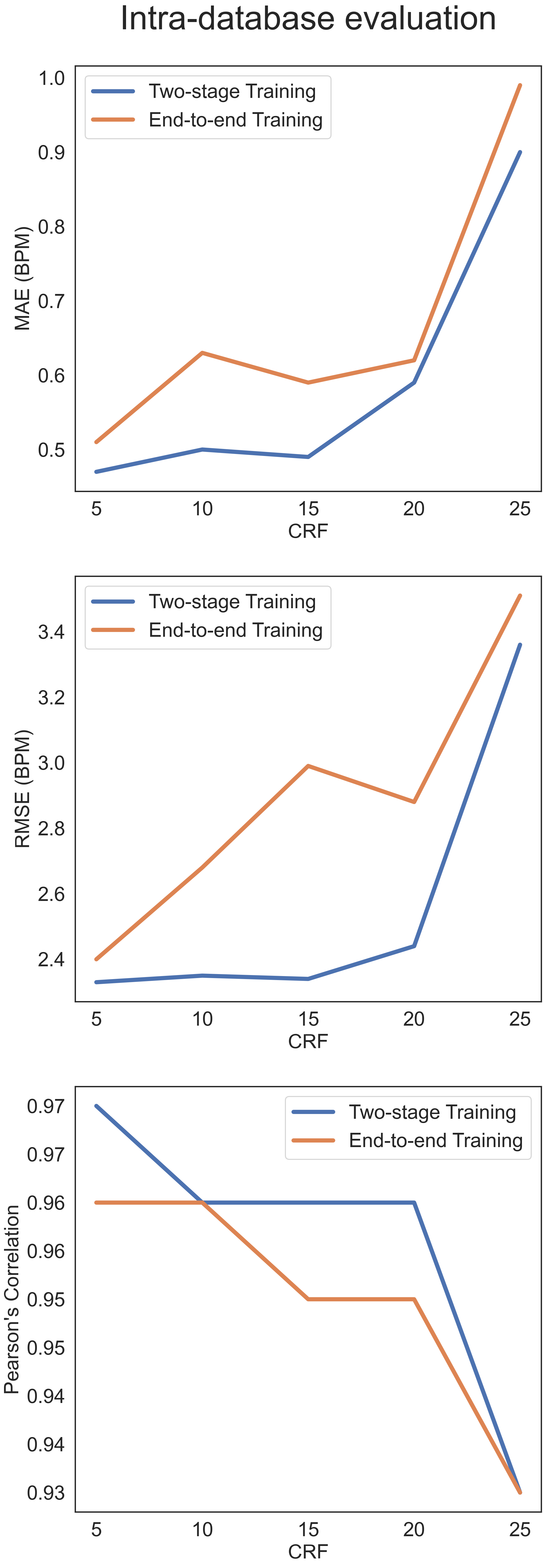}
    \includegraphics[width=3.5cm,height=9cm]{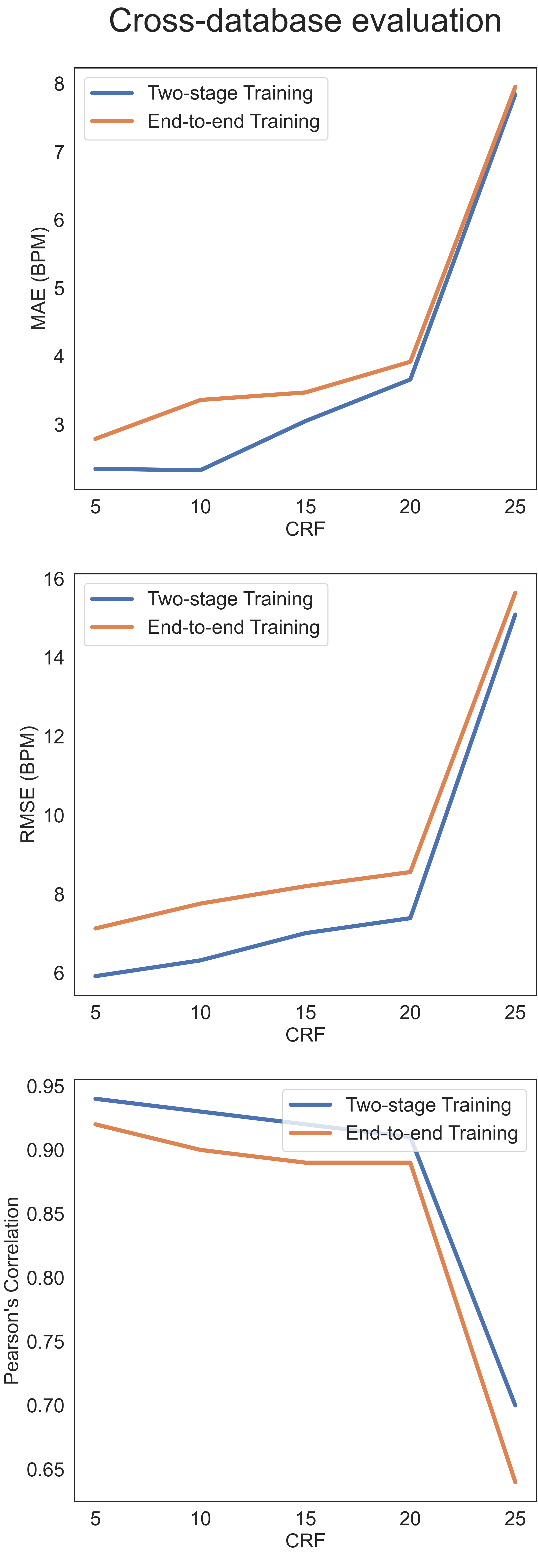}
    \caption{Training procedure evaluation for rPPG recovery in intra-dataset and cross-dataset evaluation.}
\label{fig:unet_graphics}
\end{figure}

Figure \ref{fig:unet_graphics} depicts the evolution of HR errors using both strategies for the UCLA-rPPG and UBFC-rPPG datasets as a function of the compression level. When comparing the two strategies, it is evident that both yield competitive outcomes. However, the two-stage strategy demonstrates a lower HR error in both evaluations. Specifically, within the CRF range of 0 to 20, the two-stage procedure significantly reduces HR error compared to the end-to-end framework. This reduction helps mitigate compression degradation and achieves results similar to the uncompressed scenario. On the other hand, when the CRF is set to 25, both strategies experience an increase in HR error for both intra- and cross-dataset evaluation and their results become quite similar.


The superiority of the two-stage strategy in HR estimation 
can be explained by various factors. Firstly, in the end-to-end strategy, the network aims to recover the rPPG without any constraints related to the physiological nature of the signal. This 
increases the risk of overfitting, since the system must train a larger number of parameters at the same time and the function learned by each of the two composing blocks is not necessarily the one expected by design. 
In contrast, the two-stage strategy employs a pre-trained TDM network that operates on uncompressed videos. This 
facilitates that the Pulse-Signal Magnification network (PSMN) learns a 
video transformation that focuses solely on 
enhancing 
the rPPG signal and directs attention exclusively to the regions of interest for rPPG. Hence, the initial training phase serves as a regularizer for the PSMN, ensuring pulse-signal magnification of the input data for extracting rPPG under conditions 
that match those from the
initial pre-training dataset.

\begin{figure*}[b]
    \centering
    \includegraphics[width=140mm,height=70mm]{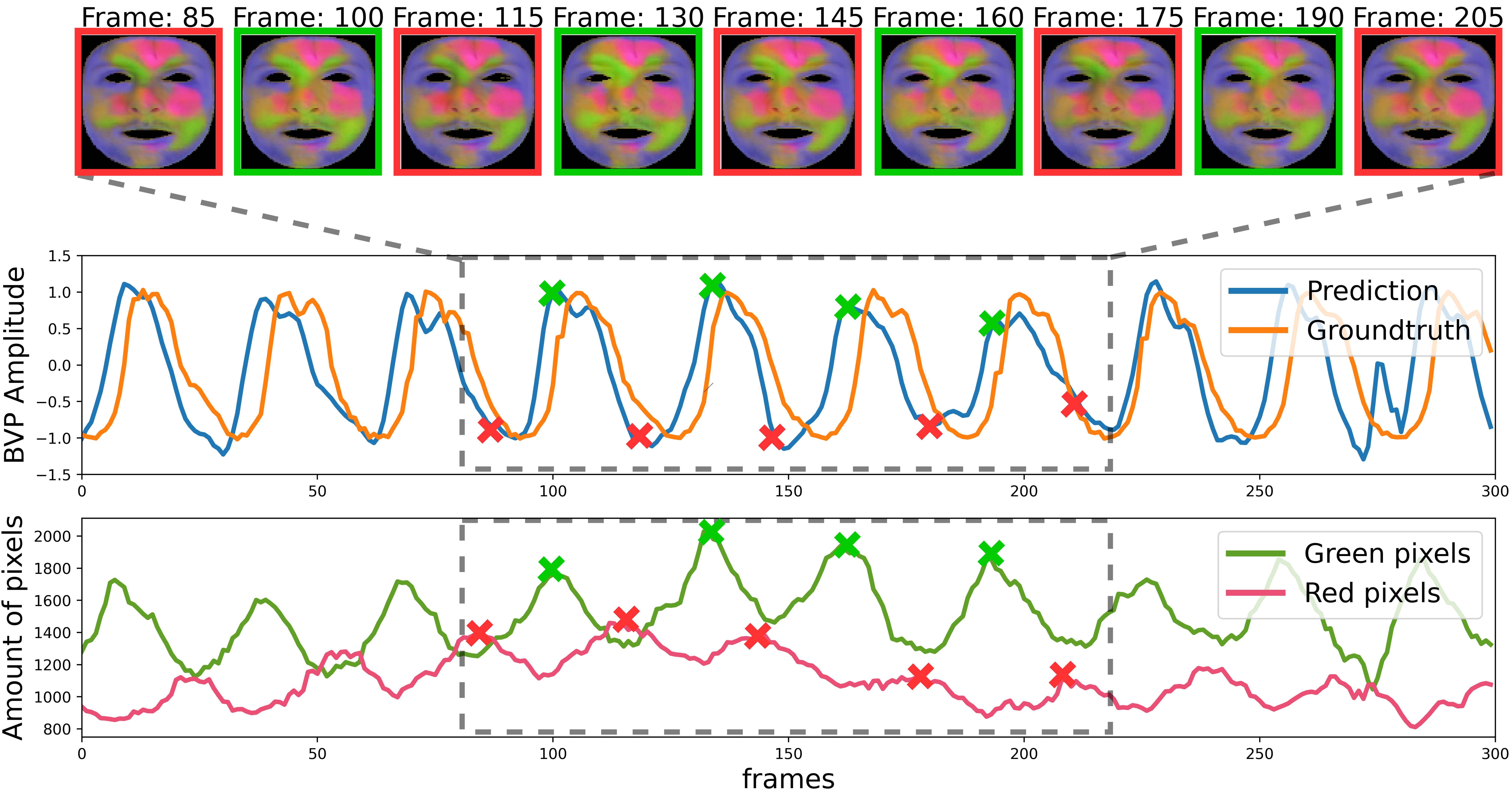}
    \caption{Visualization of the learned video transformation for a sample compressed at CRF 15 from the UCLA-rPPG dataset. The top section displays the generated 
    video, $m_\psi(\mathbf{C}^n)$
    between frames 80 and 210. In the middle, we see the predicted rPPG and ground-truth signals. The bottom section shows the number of pixels over time for green and magenta colors, which are the dominant ones after the transformation and capture the blood pulse effect (see also the  \href{https://youtu.be/bIv4itnT2mE}
    {supplementary video}).}
    \label{fig:inference}
\end{figure*}

Independently of the training strategy, we can also analyze the impact of the 
PSMN by comparing the results in Figure \ref{fig:unet_graphics} to those from Table \ref{table:derivative_impact}. Concretely, when using our PSMN, 
the error remains very similar to the uncompressed case (CRF = 0) for compressions up to CRF of 20, 
even for the cross-dataset evaluation, which was not the case when using just the TDM model. Beyond a CRF of 25, the HR error increases considerably, but it remains notably lower than the errors for the same CRF without the PSMN reported in Table \ref{table:derivative_impact}. 

To gain a deeper understanding of the behavior of our proposed framework, Figure \ref{fig:inference} provides a detailed example of the rPPG estimation using the two-stage strategy. The visualization reveals that the 
PSMN is learning a latent pulse magnification video transformation that 
highlights
the facial regions relevant to rPPG extraction. Within this learned transformation, we can identify two prominent colors: green and magenta. These colors are predominantly distributed around the cheeks and forehead regions. Additionally, both in the generated video frames and the pixel evolution graph at the bottom of the figure, we can observe that the network learns opposing pulse 
patterns using these two colors, which are correlated to the cardiac cycle. Specifically, green regions are primarily activated during the systolic phase, whereas magenta regions are predominantly present during the diastolic stage.
\\

\subsubsection{\textbf{Evaluation on highly compressed datasets}}

After evaluating the training procedure on our ablation datasets, we also examine the 
performance of our approach
on two highly compressed datasets: MAHNOB-HCI and COHFACE. Similarly to the previous sections, we use both cross-database and intra-database testing protocols.

For cross-database evaluation, we train on the UCLA-rPPG dataset compressed 
at the same rate as the test sets. 
Specifically, for the MAHNOB-HCI dataset we use the H.264 codec with a constant bit rate of 4200 kb/s while for the COHFACE dataset, we use the MPEG-4 codec with a constant bit rate of 250 kb/s. After compressing the UCLA-rPPG dataset, we 
use it to train our model and
directly test on the MAHNOB-HCI and COHFACE datasets.

For intra-database evaluation on the MAHNOB-HCI, we 
initialize our model by pre-training on
the compressed UCLA-rPPG dataset at 4200 kb/s and fine-tune it on the MAHNOB-HCI, following the subject-independent 9-fold cross-validation protocol introduced by Yu et al. \cite{yu2019remote}. Since there is no PPG signal for this dataset, we perform the fine-tuning considering just the SNR loss, which only requires HR estimates. 

For intra-database evaluation in COHFACE, we 
initialize our model by pre-training on 
the compressed UCLA-rPPG dataset at 250 kb/s and fine-tune it on the COHFACE using its standard partition protocol \cite{heusch2017reproducible}. Unlike the MAHNOB-HCI dataset, COHFACE contains PPG ground-truth data, allowing us to use the combined loss function introduced in Eq. \ref{combined_eq}.

Table \ref{table:unet_mahnob_cohface} summarizes the results. Firstly, we observe that the HR errors in the intra-dataset settings are excellent for both datasets, with the two-stage training slightly outperforming the end-to-end training on MAHNOB-HCI and more clearly in COHFACE. 
This highlights the ability of the PSMN to fine-tune the transformation of the input video to the characteristics of the targeted database. Particularly, the two-stage strategy exhibits impressive results on the COHFACE database, with 0.70 BPM (Beats Per Minute) MAE, 1.53 BPM RMSE, and 0.98 Pearson's Correlation. These findings align well with the trends depicted in Figure \ref{fig:unet_graphics}, where intra-database results demonstrate competitive even at a CRF of 25, which is approximately equivalent to a bit rate of 250 kb/s used to encode COHFACE.

On the other hand, the cross-database errors are comparatively higher, but still at the top performance reported in 
the literature under
comparable settings. For MAHNOB-HCI, 
the
two-stage training is superior to the end-to-end approach as a result of 
the regularization effect 
of the pre-trained TDM.
Nevertheless, this does not hold for the COHFACE dataset, in which neither of the strategies performs well 
under cross-database settings. We attribute this behaviour to the important differences in acquisition conditions
between the UCLA-rPPG dataset and the COHFACE dataset. 

\begin{table}[tb]
  \caption{Evaluation of PSMN training procedures on highly compressed datasets.}
  \renewcommand{\arraystretch}{1.75}
  \centering
  \adjustbox{width=0.49\textwidth}{
  \small 
  \begin{tabular}{c|c|ccc|ccc}
    \hline
    \multirow{2}{*}{Eval} & \multirow{2}{*}{Strategy} & \multicolumn{3}{c|}{MAHNOB-HCI} & \multicolumn{3}{c}{COHFACE}\\ 
    \cline{3-8}
    && MAE$\downarrow$ & RMSE$\downarrow$ & R$\uparrow$ & MAE$\downarrow$ & RMSE$\downarrow$ & R$\uparrow$\\ 
    \hline
    \multirow{2}{*}{\rotatebox{90}{\parbox{1.2cm}{\centering Cross-Database}}} & Two-stage & 3.66 & 5.25 & 0.91 & 10.42 & 17.43 & -0.07\\  
    & End-to-end & 5.57 & 7.00 & 0.90 & 10.12 & 17.19 & -0.04\\
    \hline
    \multirow{2}{*}{\rotatebox{90}{\parbox{1.2cm}{\centering Intra-Database}}} & Two-stage & 2.40 & 3.37 & 0.94  & 0.70 & 1.53 & 0.98\\  
    & End-to-end & 2.41 & 3.49 & 0.93 & 0.86 & 2.66 & 0.96\\
    \hline
  \end{tabular}
  }
  \label{table:unet_mahnob_cohface}
\end{table}

\subsection{Comparison with existing methods}
\label{subsec:comparison}
In this section we compare our results to those reported by other methods on the COHFACE and MAHNOB-HCI datasets, which are two widely used benchmarks in HR measurement with high compression rates.
\\

\subsubsection{Evaluation on MAHNOB-HCI dataset}


Table \ref{table:mahnob} gathers the results reported in the literature for this database, separating them into intra- and cross-database settings to allow for a fair comparison. In terms of cross-database evaluation, our 
method
achieves the best results, with a MAE of 3.66 BPM and an RMSE of 5.25 BPM, outperforming handcrafted methods and all deep learning approaches that report results in cross-database settings. Furthermore, our results outperform some learning-based approaches directly trained on the MAHNOB-HCI dataset, 
which
can be attributed to our two-stage training procedure
that allows the PSMN
to learn an appropriate transformation for accurate rPPG estimation. By incorporating the TDM module, we 
add constraints that prevent
the model from overfitting while effectively extracting 
meaningful
rPPG features. 
\begin{table}
\centering
\caption{Results of average HR estimation of MAHNOB-HCI.}
\renewcommand{\arraystretch}{1.3}
\centering
\adjustbox{width=0.49\textwidth}{
  \begin{tabular}{c|c c c | c c c}
    \hline
    \multirow{2}{0.7cm}{Method} & \multicolumn{3}{c}{Cross-dataset eval} &\multicolumn{3}{|c}{Intra-dataset eval}\\\cline{2-7}
    \cline{2-7}
    & MAE$\downarrow$&RMSE$\downarrow$&R$\uparrow$  
    & MAE$\downarrow$&RMSE$\downarrow$&R$\uparrow$

    \\ 
    \hline
       Poh2011 \cite{poh2010non}& - & 13.6 & 0.36 & - & - & -  \\
       CHROM \cite{de2013robust} & 13.49  & 22.36 &  0.2 & - & - & -  \\
       Li2014 \cite{li2014remote} & - &  7.62 &  0.81 &  &  &   \\
       SAMC \cite{tulyakov2016self} & 4.96 & 6.23 & 0.83 & - & - & -  \\
       HR-CN\cite{vspetlik2018visual}  & - & - & - & 7.25 &  9.24 &  0.51   \\
       DeepPhys \cite{chen2018deepphys} & 4.57 & 6.44   & 0.84 & - & - &  - \\ 
       RhythmNet \cite{niu2019rhythmnet} & - & 8.28 & 0.64 & - & 3.99 & 0.87 \\ 
       STVEN+rPPGNet \cite{yu2019remote} &  - & - & - & 4.03 & 5.93 & 0.88  \\
       STMap+CNN \cite{song2020heart} &  5.98 & 7.45 & 0.75 & 4.61  & 5.70 & 0.86 \\
       AutoHR \cite{yu2020autohr} &  - & - & - & 3.78  & 5.10 & 0.86  \\
       
       Meta-rPPG \cite{lee2020meta} & - & - & - & 3.01 & 3.68 & 0.85       \\ 
       PulseGAN\cite{song2021pulsegan} & 4.15 & 6.53 & 0.71 & - & - & -       \\ 
       rPPG-FuseNet\cite{jaiswal2022rppg} & - & - & - & \textbf{2.08} & 3.41 & 0.92       \\ 
       PhysFormer++ \cite{yu2023physformer++} & - & - & - & 3.23 & 3.88 & 0.87 \\ 
    \hline 
    \textbf{Proposed method} & \textbf{3.66} & \textbf{5.25} & \textbf{0.91} & 2.40 & \textbf{3.37} & \textbf{0.94}   \\
    
    \hline
    \end{tabular}
    }
    \label{table:mahnob}
    \end{table}

\begin{figure}[b]
    \centering
\includegraphics[width=85mm,height=45mm]{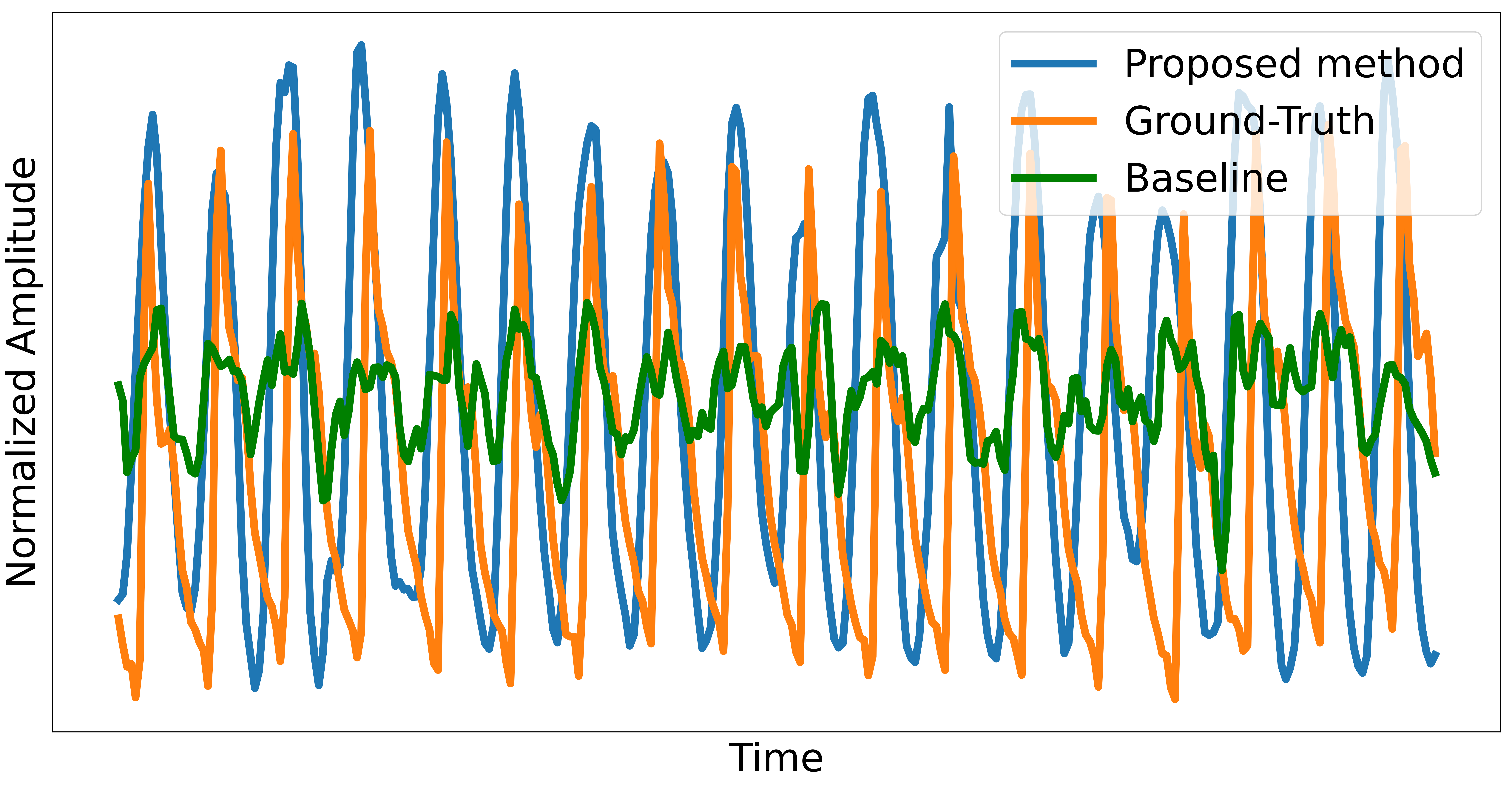}
    \caption{Comparison of rPPG estimations between our baseline (green) and proposed model (blue) in the COHFACE dataset.}
    \label{fig:cohface_comparison}
\end{figure}

Under intra-dataset settings (i.e. fine-tuning on the MAHNOB-HCI dataset), we obtain
notable improvement with respect to our cross-dataset results. The HR error 
reduces
to 2.40 BPM for MAE, 3.37 for RMSE, and 0.94 for Pearson's correlation. These 
results also outperform the current state-of-the-art approaches. 
Only the method by Jaiswal and Meenpa \cite{jaiswal2022rppg}, which incorporates a color fusion technique to eliminate illumination artifacts, achieves similar results to ours, with slightly better MAE but slightly worse RMSE and R. Interestingly, the results of STVEN+rPPGNet \cite{yu2019remote} yield significantly higher HR errors, despite of using a rather similar architecture to ours but targeting the enhancement of the entire video signal. This highlights the advantage of our approach, which focuses on restoring only the part of the video signal that is necessary to the rPPG estimation, facilitating training and improving performance.
\\

\subsubsection{Evaluation on COHFACE dataset}

The HR estimation results of the COHFACE dataset for existing methods using intra-dataset evaluation are presented in Table \ref{table:cohface}. 



Specifically, our method shows
similar HR MAE (0.70) to the best results reported to date, from the siamese-rPPG approach \cite{tsou2020siamese} but exhibits better RMSE (1.53 BPM) and R (approximately 0.98). 

The effectiveness of our proposed two-stage framework, along with the incorporation of the PSMN, is 
showcased
in Figure \ref{fig:cohface_comparison}. This figure presents a comparison between the TDM baseline model and our proposed approach, using the ground-truth PPG signal from an especially difficult 
COHFACE sample under challenging 
illumination conditions. The results highlight how the Pulse-Signal Magnification approach significantly enhances the reliability and accuracy of the rPPG signal, surpassing the performance of the TDM baseline model.

\begin{table}[t]
\renewcommand{\arraystretch}{0.95}
\centering
\caption{Results of average HR estimation of COHFACE.}
\centering
  \begin{tabular}{c|c c c}
    \hline
    \multirow{1}{1cm}{Method} & MAE$\downarrow$ & RMSE$\downarrow$ &R$\uparrow$ 
    \\
    \hline
     CHROM \cite{de2013robust}  & 7.8 & 12.45 & 0.26 \\[0.2ex]
     LiCVPR \cite{li2014remote}  & 19.98 & 25.59 & -0.44 \\ [0.2ex]
    Two stream \cite{wang2019vision} &  8.09 & 9.96 & 0.40 \\ [0.2ex]
    HR-CNN \cite{vspetlik2018visual}  & 8.1 & 10.8 & 0.29 \\ [0.2ex]
    Tsou2020 \cite{tsou2020siamese}  & 0.68 & 1.65 & 0.72  \\ [0.2ex]
    ETA-rPPGNet \cite{hu2021eta}   & 4.67 & 6.65 & 0.77       \\ [0.2ex]
    Gideon2021 \cite{gideon2021way}   & 1.5 & 4.6  & 0.90      \\ [0.2ex]
    TFA-PFE \cite{li2023learning}   & 1.31 & 3.92  & -      \\ [0.2ex]
    TDM \cite{comas2022efficient} & 2.48 & 5.90 & 0.81       \\ [0.2ex]
    \hline
    \textbf{Proposed method} & \textbf{0.70} & \textbf{1.53} & \textbf{0.98}        \\ [0.2ex]
    \hline
    \end{tabular}
    \label{table:cohface}
    \end{table}


%% file: Conclusions/Conclusions.tex
\section{Conclusions}
\label{sec:conclusions}

In this paper, we present a novel two-stage framework designed to mitigate the impact of video compression on remote HR measurement by 
enhancing and magnifying the video information required to estimate the rPPG signal.
Our model consists of two sub-networks: an rPPG estimator and a Pulse-Signal Magnification network (PSMN). 
We present extensive evaluation on 4 public datasets and show that:
\begin{enumerate}
    \item HR estimation from rPPG signals is strongly affected by compression. This is in line with numerous prior studies. We quantify the impact of various compression rates on the selected baseline model and datasets (Section \ref{subsubsec:impact}).
    \item The introduction of the PSMN has a strong impact on HR estimation error, which is now robust against compression rates up to CRF = 20. Stronger compression levels do increase the HR estimation error, although considerably less than the baseline.
    \item The proposed two-stage training outperforms the end-to-end alternative by a considerable margin, thanks to the regularization effect obtained by the rPPG estimator pre-training on uncompressed data (Section \ref{subsubsec:compression_unet_ablation}).
    \item The proposed optimization, exclusively from the rPPG perspective, yields a novel physiologically transformed video that amplifies the pulse signal, which would otherwise be imperceptible to the naked eye in the input RGB domain.
    \item The proposed model can successfully adapt to datasets that are highly compressed and recorded in conditions that are quite different to those in the uncompressed pre-training data. The obtained results on the two most popular public datasets in such category show top state-of-the-art performance.    
\end{enumerate}

%% file: Acknowledgment/Acknowledgment.tex
\section*{Acknowledgments}
This work is partly supported by the eSCANFace project (PID2020-114083GB-I00) funded by the Spanish Ministry of Science and Innovation.